\documentclass[10pt]{article}
\usepackage[utf8]{inputenc}
\usepackage[T1]{fontenc}
\usepackage{amsmath}
\usepackage{amsthm}
\usepackage{amsfonts}
\usepackage{amssymb}
\usepackage{makeidx}
\usepackage{graphicx}
\usepackage{algorithm2e}
\usepackage{lmodern}
\usepackage{mathtools}
\usepackage{url}
\usepackage{xcolor}
\usepackage[left=2cm,right=2cm,top=1cm,bottom=2cm]{geometry}
\usepackage{multirow}
\usepackage{caption}
\usepackage{subcaption}
\usepackage{tabularx} 
\usepackage{booktabs}

\author{Prudence Djagba\thanks{Michigan State University, USA. 
		\href{mailto:prudence.djagba@aims.ac.rw}
{djagbapr@msu.edu}} 
	\and  J. K. Buwa Mbouobda \thanks{AIMS Ghana, Accra. 
		\href{mailto:arame.sow@aims-senegal.org}{ jordan@aims.edu.gh}}}
\title{Deep Transfer Learning for Breast Cancer Classification}

\usepackage{hyperref}
\usepackage{cleveref}

\theoremstyle{remark}

\theoremstyle{definition}

\date{}
\begin{document}
\maketitle	
\begin{abstract} 
\noindent

Breast cancer is a major global health issue that affects millions of women worldwide. Classification of breast cancer as early and accurately as possible is crucial for effective treatment and enhanced patient outcomes. Deep transfer learning has emerged as a promising technique for improving breast cancer classification by utilizing pre-trained models and transferring knowledge across related tasks. In this study, we examine the use of a VGG, Vision Transformers (ViT) and Resnet to classify images for Invasive Ductal Carcinoma (IDC) cancer and make a comparative analysis of the algorithms. The result shows a great advantage of Resnet-34 with an accuracy of $90.40\%$ in classifying cancer images. However, the pretrained VGG-16 demonstrates a higher F1-score because there is less parameters to update. We believe that the field of breast cancer diagnosis stands to benefit greatly from the use of deep transfer learning. Transfer learning may assist to increase the accuracy and accessibility of breast cancer screening by allowing deep learning models to be trained with little data. \\

\noindent
\textbf{Keywords:} Breast cancer, Deep Transfer Learning, Vision Transformers.
\end{abstract}
 \section{Introduction}

 According to GLOBOCAN 2020, there were 10.3 million cancer-related deaths and 19.3 million new cases of the disease worldwide. Female breast cancer represented 11.7\% of all cases, or 2.26 million cases, globally \cite{chhikara2023global,GCO}. In Africa, year 2020, $85\,787$ women died from breast cancer out of $186\,598$ new cases, $12.5\%$ of world's death by breast cancer \cite{GCO}. In the same year, we diagnosed $2\,262\,419$ new cases of breast cancer in the world \cite{GCO}. Through the use of computer-aided procedures, the accuracy of diagnosis may be significantly boosted. One of the most efficient methods to detect breast cancer is a pathologist's examination of histopathological pictures under a microscope \cite{budak2019computer, das2021breast}. Since the invention of digital image scanners, picture identification has improved \cite{alghodhaifi2019predicting}. The practice of digital pathology has advanced significantly, in part because whole slide image (WSI) scanners enable quick and affordable diagnosis \cite{bolhasani2020histopathological}. Breast cancer of the Invasive Ductal Carcinoma (IDC) kind comes from a dysfunction of  milk duct's cells of the breast and eventually spreads to the breast tissue nearby. It is an example of cancer type where early detection and accurate classification can significantly improve patient outcomes. The results for patients can be considerably improved by IDC early identification and precise categorization. On the basis of histopathological scans, IDCs have been accurately classified using machine learning approaches \cite{abdolahi2020artificial, pratiwi2021invasive}. Diagnosing pathological imaging is difficult and necessitates manual skills and a microscope. Due to the limitations of human cognition, this time-consuming approach frequently results in mistakes. One might anticipate time savings and decreased mistake rates with an automated system \cite{pratiwi2021invasive}.

 One of the main challenges in analyzing cancer images using deep learning algorithms is the limited availability of data \cite{diagnostics13010161,Singh2023}. Deep learning algorithms require large amounts of data to train effectively, and the lack of data can lead to overfitting or underfitting of the model \cite{diagnostics13010161}.
 Many models have been created to detect malignant cells for quite some time, and deep learning models have been used in that regard. \cite{BARSHA2021104931}. But, one can ask that: can deep learning algorithms accurately predict the presence of cancer tumor in image data? In this paper, we will use Vision Transformers, VGG, and ResNets to classify images of breast cancer. Therefore in this paper we aims to: Give a useful and clear overview of neural networks and how they work;  Describe some deep learning algorithms such as convolutional neural networks, VGGnets, residual networks and vision transformers; Train the different models on the invasive ductal carcimona data set and  Evaluate each of the model and compare their accuracy.

This paper  contains three sections. In  section 2 we  discuss about specific architectures that will be used in this work and describe their functioning. Next, in section 3 we present the result of our experiments on the different networks and then compare the results. Finally, in section 4, we give a clear summary of what has been done in the work and give perspectives for futures works.

 \section{Methodology}
This section describes the methodology employed to analyze the cancer dataset. It concentrates on convolutional neural network (CNN) architectures like ResNet, VGG in addition to the emerging field of vision transformers. This chapter provides an overview of the dataset, discusses the preprocessing steps, and introduces the selected CNN architectures and their potential cancer classification applications. In addition, the use of vision transformers as an alternative method is highlighted.

\subsection{Dataset description}
\subsubsection{Data source}

We use the dataset provided at \url{https://bit.ly/3XBOdum}.

\begin{figure}[h]
	\centering
	\includegraphics[width=0.6\linewidth]{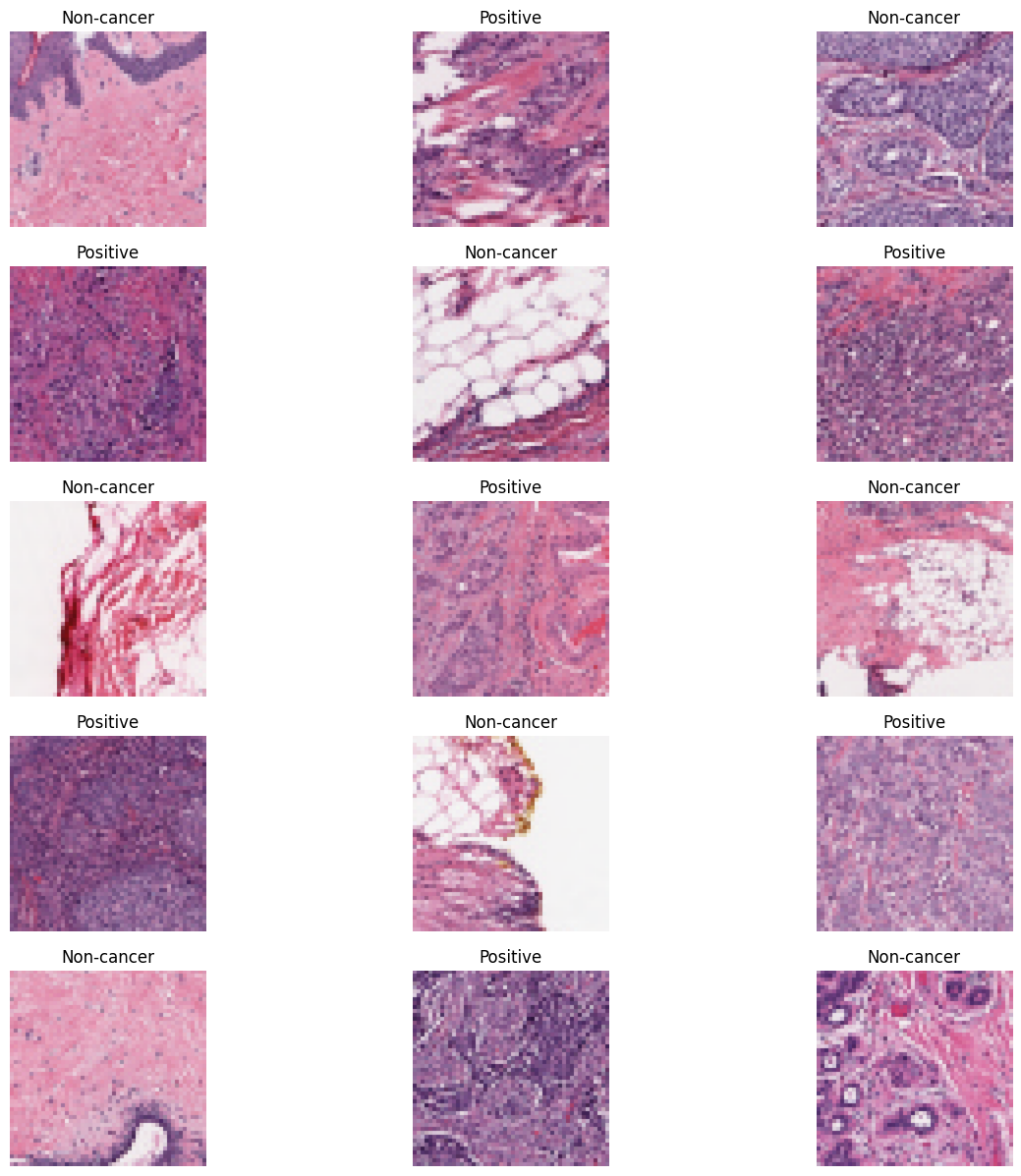}
	\caption[Some example of images in the breast histopathology imaging dataset for both courses.]{Some example of images in the breast histopathology imaging dataset for both classes.}
	\label{fig:cancerimg}
\end{figure}

The majority of occurrences of breast cancer are caused by invasive ductal carcinoma (IDC). When evaluating a specimen's aggressiveness as it is put on the test slide of the microscope, pathologists pay special attention to the areas that contain the IDC. Delineating the specific IDC zones inside a whole-mount microscope slide is therefore a typical preprocessing step for automated aggressiveness evaluation. The original dataset consisted of 162 whole-mount slide images of Breast Cancer (BCa), from women patients at the Hospital of the University of Pennsylvania and The Cancer Institute of New Jersey \cite{abdolahi2020artificial}. The original microscope slides used to create these  photos of breast cancer (BCa) were magnified at a level of 40. They are separated into 277,524 patches, each 50 pixels by 50 pixels in size. The IDC-positive class sample count in this total is 78,786; the IDC-negative class sample count is 198,738. Figure \ref{fig:cancerimg} presents the dataset's sample pictures.

\subsubsection{Image preprocessing}

Both the identification of breast cancer and the accuracy with which a model makes predictions might be hampered by the presence of artifacts and noise. As a result, it is essential to process the images in order to enhance the performance of a model. The aim of pre-processing is an improvement of the image
data that suppresses unwilling distortions or enhances some image features
important for further processing, although geometric transformations of images (e.g. rotation, scaling, translation) are classified among pre-processing
methods here since similar techniques are used \cite{sonka2014image}.

Image augmentation is a method that is used in machine learning and computer vision to artificially extend a dataset by creating new modified versions of images.  It is especially helpful when the original dataset has a limited number of samples since it helps minimize overfitting and enhances the performance of deep neural networks.

Here are some important details about the enhancement of images:

\begin{itemize}
	\item \textbf{Aim}: The most important aim of picture augmentation is to enhance the diversity of the dataset. This will enable the model to learn from a greater variety of variations and will improve its capacity to generalize to new data.

\item \textbf{Techniques}: Image augmentation entails making numerous modifications to the source photos, such as rotating, scaling, cropping, flipping, and adjusting the brightness, among other things. It is possible to build enhanced pictures by combining these alterations or applying them in a random order.

\item \textbf{Implementation}: Image augmentation can be accomplished via the use of libraries or frameworks that offer built-in functions or application programming interfaces (APIs) for the purpose of creating augmented images. For the purpose of doing picture augmentation, for instance, the ImageDataGenerator class is often used in deep learning frameworks such as Keras. In this work, we use enhancement methods from Pytorch and Keras frameworks.

\item \textbf{Benefit}: Image augmentation helps to expand the size of the training dataset, which may lead to improved model performance and generalization. Benefits Image augmentation helps to increase the size of the training dataset, which can lead to improvements in model performance and generalization. In addition to this, it helps to avoid overfitting by injecting variances in the data, which in turn makes the model more resistant to a variety of settings and scenarios.

\end{itemize}
Picture augmentation is a strong approach that allows for the growth and diversity of picture datasets. This improves the performance and generalization capacity of machine learning models, especially in computer vision applications. In general, image augmentation is a technique that allows for the expansion and diversification of image datasets.

\subsubsection{Dataset splitting}\hfill

It is standard procedure to separate the dataset into three separate sets when working with machine learning models \cite{refaeilzadeh2009cross}. These sets are referred to as the training set, the testing set, and the validation set respectively. The following is a description of each set, along with the benefits that come with them:

\begin{itemize}
	\item  The \textbf{training set} is the biggest part of the dataset and is what is utilized to train the machine learning model. It is the set that the model uses to discover the underlying connections and patterns in the data that it is working with. A training set offers the following benefits to its users:

\begin{itemize}
	\item  \textbf{Model learning}: The training set enables the model to learn from the data and alter its parameters in order to reduce the amount of mistake or loss function.

\item  \textbf{Pattern identification}: The model is able to discover and capture the underlying patterns and connections in the data if it is trained on a large and varied training set, which is required for the process.

\item  \textbf{Parameter optimization}: The training set is used in the process of optimizing the model's parameters by using methods such as gradient descent or backpropagation.
\end{itemize}

\item  \textbf{Testing set}: The purpose of the testing set is to assess the performance of the trained model on data that it has not seen before. It is helpful in determining how effectively the model generalizes to new cases that have not been observed before. The following is a list of the benefits of employing a testing set:

\begin{itemize}
	\item  \textbf{Performance evaluation}: The testing set provides an objective assessment of the performance of the model when applied to fresh data. It is useful in determining the extent to which the model can generalize and provide reliable predictions.
 
\item  \textbf{Overfitting detection}: By analyzing the model using the testing set, we may determine whether or not the model has overfit the training data.  When a model performs well on the data it was trained on but badly on fresh data, this is an example of overfitting.

\item  \textbf{Model comparison}: The testing set enables the comparison of several models or algorithms in order to determine which one performs the best.
\end{itemize}

\item  \textbf{Validation set}: The validation set is used in the process of adjusting hyperparameters and selecting models. It does this by fine-tuning the hyperparameters and choosing the model that performs the best overall. This enhances the overall performance of the model. Using a validation set comes with a number of benefits, including the following:

\begin{itemize}
	\item  \textbf{Hyperparameter Tuning}: The validation set is used to analyze several possible combinations of hyperparameters in order to pick the model that performs the best. It contributes to the optimization of the model's performance as well as its capacity for generalization.

\item  \textbf{Preventing Overfitting}: If we use a separate validation set, we can avoid overfitting to the testing set and guarantee that the performance of the model is not skewed toward a particular set of hyperparameters. This is possible because we can prevent overfitting to the testing set.

\item  \textbf{Model Selection}: The validation set enables the comparison of many models or architectures in order to pick the one that delivers the highest level of performance.
\end{itemize}

\end{itemize}
In a nutshell, one of the most important steps in machine learning is to segment the dataset into training, testing, and validation sets. The validation set is used for hyperparameter tweaking and model selection, while the training set is used to train the model. The testing set is used to assess how well the model performs on data that it has not seen before. This strategy helps ensuring that the model works well on fresh data, prevents overfitting, and enables the selection of the best possible model to use.

\subsection{Convolutional neural networks}\label{key}

\begin{figure}[!h]
	\centering
	\includegraphics[width=\linewidth]{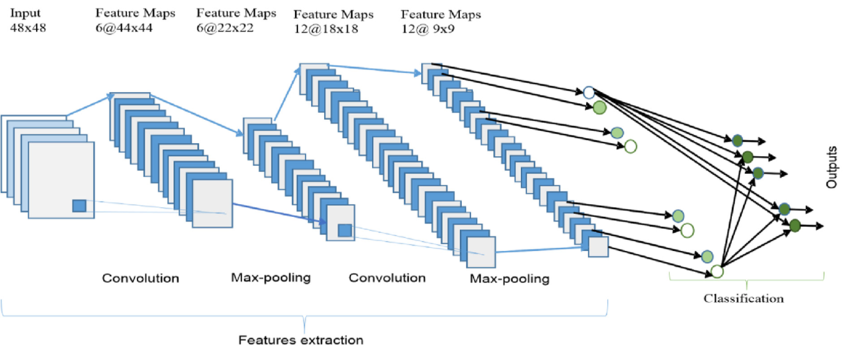}
	\caption[Typical architecture of a CNN]{Typical architecture of a convolutional neural network \cite{article}.}
	\label{fig:cnnarchitecture}
\end{figure}

Convolutional Neural Network (CNN) is a deep network that replicates how the human visual cortex processes and recognizes pictures \cite{phil2017matlab}. The Figure \ref{fig:cnnarchitecture} below is a graphical representation of a CNN. From the figure, CNN's architecture is made up of neural layers that aid in feature extraction and additional neural layers that serve as classifiers. Convolution layer, pooling layer, activation function, and fully connected layer are the minimum number of components that make up the CNN model \cite{indolia2018conceptual}. The feature extraction part consists of  convolutional and pooling layers, while the fully connected part is the classifier.  A feature map is the output picture of a convolutional layer \cite{phil2017matlab}. The convolutional layer (CL) is the main part of a CNN \cite{li2014medical}. In the CL, we use a filter to extract features (or patterns) in the input image. The input image is defined as a three dimensional matrix {height*width*depth}. The shape {height*width} is the pixel size of the image and the {\textit{depth}} represents the number of channel (3 for colored pictures (RGB) and 2 for gray-scale images). The input image is then convoluted with the kernel (filter), which results to the feature map. The convolution equation for each input image is given by \cite{li2014medical}:
\begin{equation}\label{conv}
	Feature\_map_j = \sum_{i=1}^{N}I_i*K_{i,j}+B_j,
\end{equation}
where the $I_j$ represent the input matrices, $K_{i,j}$ the corresponding kernels and $B_j$, the biases.

% TODO: \usepackage{graphicx} required

The pooling layer helps to reduce the dimension of the input. It chooses the most relevant features from the feature map and passes the result to the activation layer. The activation layer introduces the non-linearity in the model. They are many activation functions. Some of them are: the sigmoid function, the softmax function, the hyperbolic tangent function and the Rectified Linear Unit(ReLU) function. The equations are given below.
\begin{equation}\label{sigmoid}
	(\text{Sigmoid})\ \ \ \sigma(x) = \frac{1}{1+e^{-x}},
\end{equation}
\begin{equation}\label{hyptan}
	(\text{Hyperbolic tangent})\ \ \ tanh(x) = \frac{e^x-e^{-x}}{e^x+e^{-x}},
\end{equation}
\begin{equation}\label{relu}
	(\text{ReLU})\ \ \ \ ReLU(x) = \max(0,x) = \left\{\begin{array}{ccc}
		0 &\text{ if }& x<0,\\
		x &\text{ if }& x\ge0.
	\end{array}	\right.
\end{equation}
\begin{equation}\label{softmax}
(\text{Softmax})\ \ \ \ \sigma(\mathbf{z})_i = \frac{e^{z_i}}{\sum_{j=1}^K e^{z_j}} \ \ \text{ for } i = 1, \dotsc, K \text{ and } \mathbf{z} = (z_1, \dotsc, z_K) \in \mathbb{R}^K.
\end{equation}
Recent studies pointed the ReLU function to be a good activation function for classification problems using CNN as its improves both the learning speed and the performance of the models \cite{krizhevsky2012imagenet}. In this work we will use a ReLU activation function. The next part is a set of fully connected layers (a basic neural network) which helps for classification.

\subsection{The VGG architecture}

Deepchecks \cite{vgg4} describes the VGG architecture as a convolutional neural network (CNN) architecture that has been in use for some time. It was created by the Visual Geometry Group (VGG) at the University of Oxford as a consequence of research on how to make certain networks denser \cite{vgg1}. The network is distinguished by its simplicity, with the only additional components being pooling layers and a completely linked layer \cite{vgg2}. The network makes use of modest filters measuring 3 by 3, and it is also known for its compact size. The architecture is a conventional deep CNN architecture with numerous layers, and the term "deep" refers to the number of layers with VGG-16 or VGG-19 consisting of 16 or 19 convolutional layers, respectively \cite{vgg1,vgg3}. 

The network receives an image with the dimensions (224, 224, 3) as its input \cite{vgg6}. The first two layers have the same padding and each feature 64 channels with a filter size of 3 by 3. Following that, a max pool layer with a stride of (2, 2) is followed by two convolution layers with a filter size of 128 and (3, 3). After this comes a max-pooling layer with the stride (2, 2), which is the same as the layer before it. After that, there are two convolution layers with a filter size of (3, 3) and 256  filters \cite{vgg6}. 

One of the most widely used designs for picture recognition is called the VGG architecture, and it serves as the foundation for models of object identification that have broken new ground \cite{vgg1}. In addition, VGGNet performs better than the baselines on a wide variety of tasks and datasets, not only ImageNet \cite{vgg6}. All of the hidden layers in VGG employ ReLU, which considerably cuts down on the amount of time required for training. On the other hand, Local Response Normalization (LRN) is not used by VGG since it both increases the amount of memory consumed and the amount of time required for training without enhancing accuracy \cite{vgg4}. 

\begin{figure}[h!]
	\centering
	\includegraphics[width=0.7\linewidth]{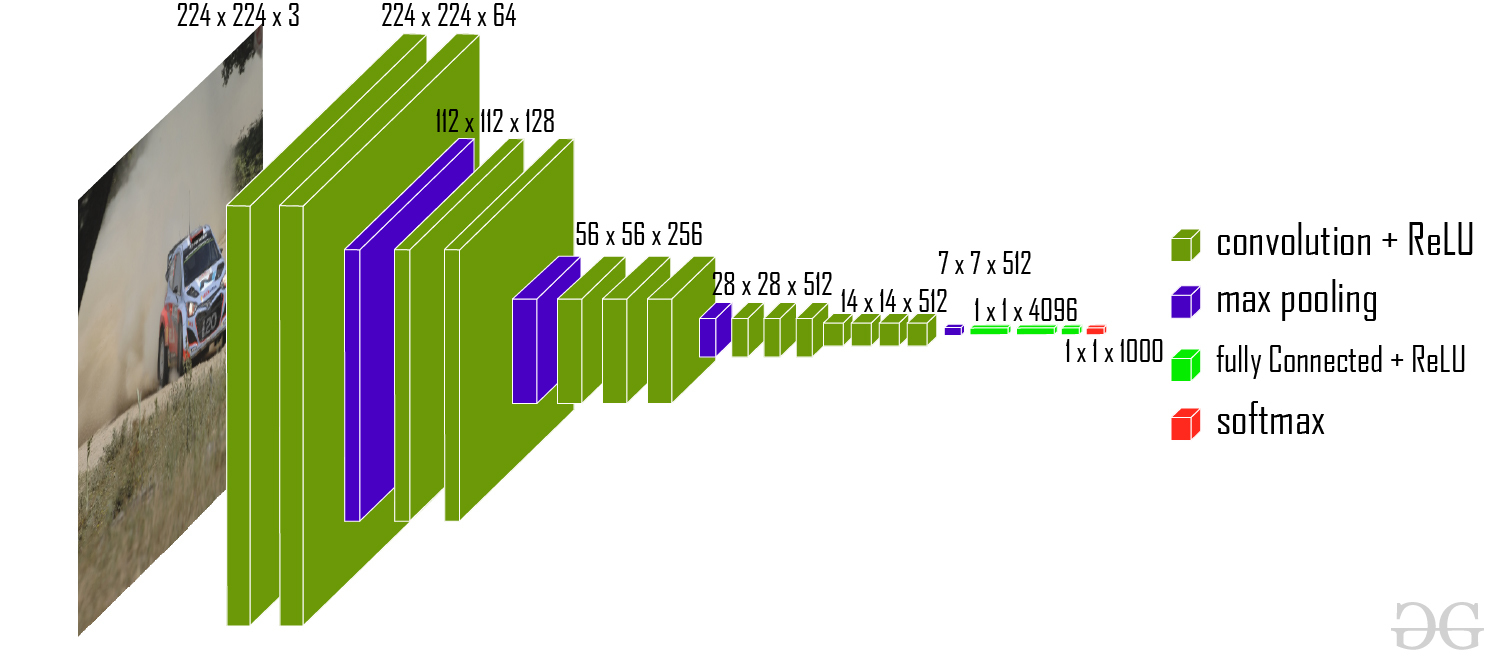}
	\caption[The VGG16 architecture]{The VGG16 architecture \cite{vgg6}}
	\label{fig:vgg16}
\end{figure}

The architecture which will be used in this essay will be a VGG16. The figure \ref{fig:vgg16} is a diagram of the network.
% TODO: \usepackage{graphicx} required

\subsection{Residual networks (ResNet)}
It has became difficult to talk about CNN architectures without mentioning residual networks. Residual Networks (abbreviated as ResNet) were originated by  \cite{he2016deep} in 2015 and were afterwards launched by Microsoft Research, winning many awards. 

\begin{figure}[!h]
\centering
	\includegraphics[width=0.7\linewidth]{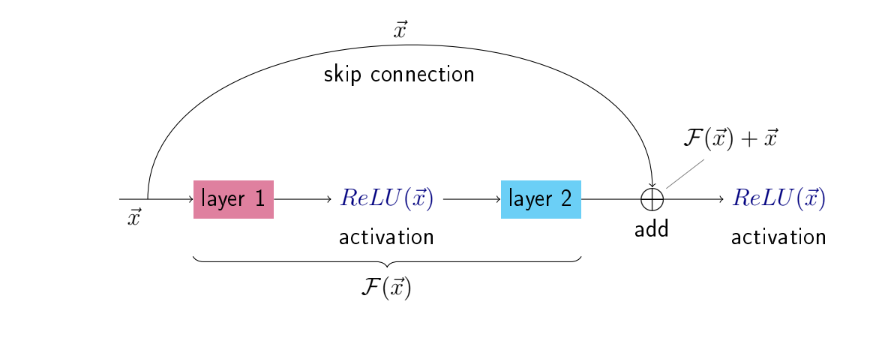}
\caption{A residual block in the ResNet architecture}
\label{kenya}
\end{figure}

The idea came out to solve the problem of vanishing or exploding gradient when it comes to build deep neural networks. Resnet are built with residual connections, forming residual blocks (Figure \ref{kenya}), which allows the information (the learning process) to navigate through the network and prevent gradient from dissipating \cite{UNAIDS}. The residual connections here refers to a type of link that bypasses one or more layers in the network and reach the output directly. This shortcut connection enables the network to learn fast and to achieve excellent performances. 

There are many ResNet models depending on how the residual blocks are stacked together. The following table present some of the architectures.
\begin{table}[h!]
	\includegraphics[scale=.26]{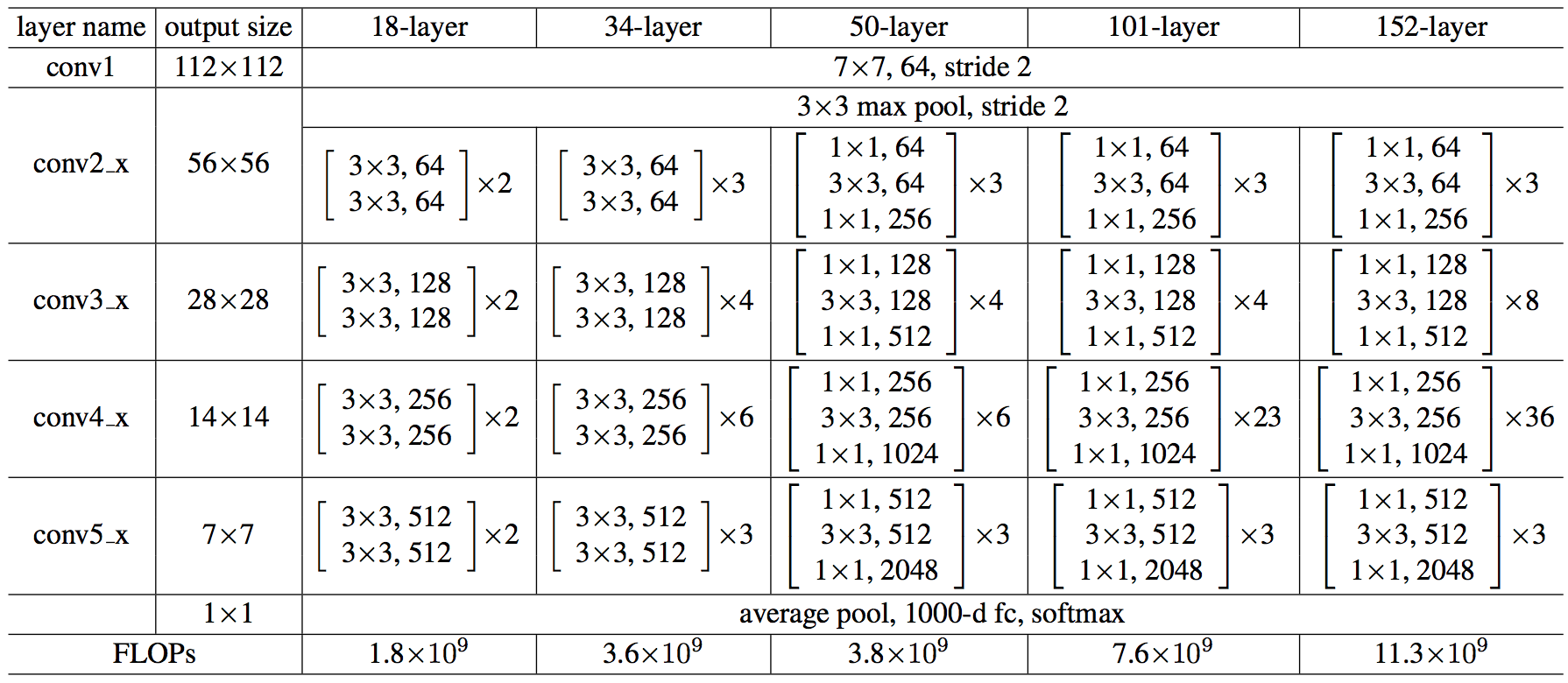}
	\caption{Some ResNet architectures \cite{he2016deep}.}\label{Resn}
\end{table}

\section{Vision Transformer for image classification}\label{vit}

Transformers is a deep learning algorithm that relies on \textbf{attention} \cite{vaswani2017attention}. It is known for his high performance in Natural Language Processing tasks. While CNN-based algorithms were seen as the best models for computer vision, vision transformer has shown that if the attention applied in NLP is applied to patches of images, it can perform very well on image classification tasks \cite{dosovitskiy2020image}. Vision transformers, when pretrained on very large image datasets (ImageNet, CIFAR-100, JFT-300M, etc...) and fine-tuned to the target data, performs very well on image recognition tasks compare to ResNet and other CNN models \cite{dosovitskiy2020image}. 

% TODO: \usepackage{graphicx} required
\begin{figure}[h!]
	\centering
	\includegraphics[scale=0.3]{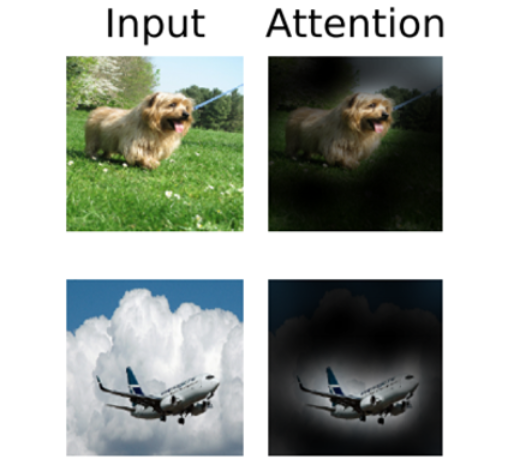}
	\caption[Example of attention on image input]{Example of attention on image input \cite{ViT}.}
	\label{fig:attention}
\end{figure}
The attention mechanism is a dot product operation between two image feature. The vision transformer is made of many blocks. Each of them plays a crucial role in processing the image.

\subsubsection{Patches creation}\hfill

At the very first step of the vision transformer algorithm the image is split into patches of size $P$. For example an image of size $320 \times 240$, can be split into $300$ patches of size $16\times16$. In the Figure \ref{fig:imagetopatch0} below, our images are of size $72\times72$ and we create $144$ patches of size $6\times6$. 

% TODO: \usepackage{graphicx} required
\begin{figure}[!h]
	\centering
	\includegraphics[width=0.3\linewidth]{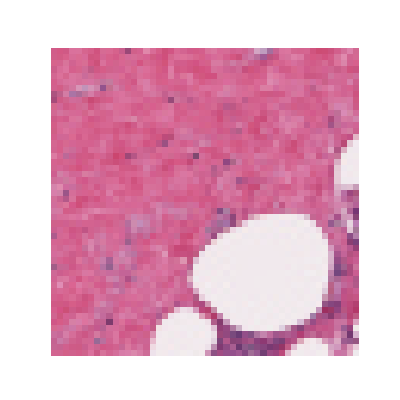}
	\includegraphics[width=0.3\linewidth]{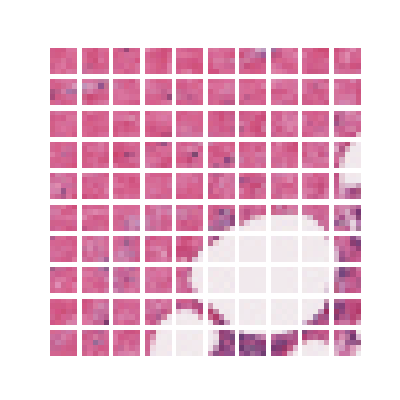}
	\caption[Patches creation]{Patches creation}
	\label{fig:imagetopatch0}
\end{figure}

 \subsubsection{Linear projection of patches}\hfill
 
 The different patches are now flattened into a vector by projection. The number of patches $N$ is the vector length. Let $\mathbf{x}\in\mathbb{R}^{N\times P^2 C}$ ($C$ is the number of channels) an image, a patch can be represented by $\mathbf{x}^i_P\in\mathbb{R}^{P\times P\times C}$. We reshape first the patches.
 \[\mathbf{x}^i_P\in\mathbb{R}^{P\times P\times C} \longrightarrow \mathbf{x}^i_P\in\mathbb{R}^{1\times P^2 C},\] Next, we compute the projection operation
 \[\mathbf{x}^i_P\in\mathbb{R}^{1\times P^2 C}\cdot \mathbf{W}\in \mathbb{R}^{ P^2 C\times D}\longrightarrow \mathbf{x}_P^i\cdot\mathbf{W}\in \mathbb{R}^{1\times D}.	\] Here, $D$ represent the target dimension of the projection.
 
  The operation can be written directly as
 \[	\mathbf{x}\in\mathbb{R}^{N\times P^2 C}\cdot\mathbf{W}\in\mathbb{R}^{P^2C\times D}=\mathbf{x}\cdot\mathbf{W}\in\mathbb{R}^{N\times D}.\]
The final result is called the patch embedding \cite{dosovitskiy2020image}.

\subsubsection{Position embedding}\hfill

The position embedding are added to the patch embedding to save the position information of the patches. They are various positions embedding method including sinusoidal function  used in the original Transformers paper \cite{vaswani2017attention} and Rotary positional embedding (RoPE), well-suited for NLP tasks.

\subsubsection{Transformer encoder}\hfill

This part constitute the main block of the transformers architecture. It helps to compute the attention. The Transformer encoder \cite{vaswani2017attention} is built with alternating layers of multiheaded self-attention  and Multi-Layer Perceptron (MLP) blocks . We apply normalize the input  before
every block, and residual connections after every block \cite{baevski2018adaptive, wang2019learning}.

The attention is computed using the formula
\begin{equation}\label{attention}
	\mathbf{Attention(Q,K,V)} = \mathbf{softmax}\left(\frac{\mathbf{QK^T}}{\sqrt{\mathbf{d_K}}}\right)V,
\end{equation}

where $\mathbf{Q}$ (query) represent the features of interest, $\mathbf{K}$ (key) are the relevant features, $\mathbf{V}$ (value) is the original feature and $\mathbf{d_K}$ is a normalization factor.

% TODO: \usepackage{graphicx} required
\begin{figure}[!h!]
	\centering
	\includegraphics[width=0.6\linewidth]{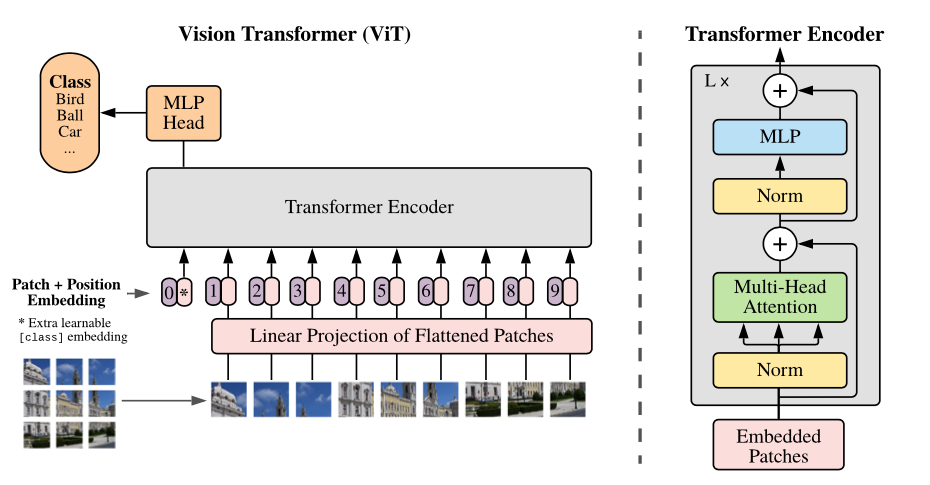}
	\caption[The overall vision transformer architecture]{The overall vision transformer architecture \cite{dosovitskiy2020image}.}
	\label{fig:vitfigure}
\end{figure}
The normalization reduce the training time, and stabilize the training while the skip of connection improve the performance. 

\subsubsection{Prediction module}\hfill

The final part of the vision transformer serves to classification. It is made of dense layers like in CNN. It is a set of MLP helping to categorize images. 

\subsection{Model evaluation}
When it comes to the evaluation of a model that uses machine learning, there are a variety of methods and metrics that may be employed. We present here some important aspects to consider while evaluating models.

\subsubsection{Accuracy}\hfill

 Accuracy is a measurement that determines how right the model's predictions are in general. It is determined by determining the ratio of the number of accurate forecasts to the total number of predictions that were made. On the other hand, accuracy by itself could not be enough in situations in which the dataset is unbalanced or in which the effects of several kinds of mistakes vary greatly from one another.
 \begin{equation}\label{accuracy}
 	Accuracy = \frac{\text{TP}+\text{TN}}{\text{TP}+\text{TN}+\text{FP}+\text{FN}}=\frac{\text{Correct predictions}}{\text{All prediction instances}}.
 \end{equation}

\subsubsection{Precision}\hfill

 Precision is defined as the ratio of accurately predicted positive occurrences to the total number of cases that were anticipated to be positive. It focuses on the capability of the model to prevent producing false positives. The ratio of the number of true positives to the total number of true positives and false positives is the formula for calculating precision;
 \begin{equation}\label{precision}
 	Precision = \frac{\text{TP}}{\text{TP}+\text{FP}} = \frac{\text{True positive}}{\text{Total positive predictions}}.
 \end{equation}

\subsubsection{Recall}\hfill

 Recall, which also goes by the names sensitivity and true positive rate, is a measurement that determines the percentage of properly predicted positive cases out of all actual positive instances. It places an emphasis on the model's capacity to recognize all instances of good outcomes. The ratio of the number of true positives to the total number of true positives and false negatives is how recall is determined;
 \begin{equation}\label{recall}
 	Recall = \frac{\text{TP}}{\text{TP}+\text{FN}} = \frac{\text{True positive}}{\text{Total positive observations}}.
 \end{equation}

\subsubsection{The F1-score}\hfill

The F1-score is a statistic that combines precision and recall into a single number. This value is referred known as the F1-score. Due to the fact that it takes into consideration both false positives and false negatives, it offers a more accurate measurement of the performance of the model. The F1-score is determined by taking the harmonic mean of the recall and precision scores;
\begin{equation}\label{f1-score}
F1\text{-}score = 2\times\frac{Recall\times Precision}{Recall + Precision}.	
\end{equation}

\subsubsection{Confusion matrix}\hfill

 A confusion matrix is table that describes the performance of a classification model is referred to as a confusion matrix. It presents a comprehensive analysis of the model's predictions, including the model's true positives, true negatives, false positives, and false negatives. It is helpful for comprehending the various kinds of mistakes that the model is producing.

In order to have a thorough comprehension of the performance of the model, it is essential to take into account a variety of assessment metrics and approaches. The particular job, dataset, and objectives of the project all play a role in determining which assessment technique to use. We are able to arrive at well-informed conclusions and increase the model's performance if we give great consideration to its evaluation.

\subsection{Transfer learning and Fine-tuning}\label{fine}

\subsubsection{Transfer learning}\hfill

Transfer learning is the process of applying deep learning models that have been trained on a big dataset to a smaller dataset in order to solve a different issue. This is accomplished by transferring the weights that were learned during a prior training phase to a new training session. The VGG-16 model that is used in this study has undergone preliminary training using the 1.2 million pictures that are included in the ImageNet dataset. It is possible to use the weights that were learned during the training phase to a different issue in order to reduce the amount of time and money spent on calculation. The transfer learning strategy will be used on the breast cancer dataset, and the pre-trained model will serve as the basis for its application. We have a number of different options available to us when it comes to how we put the transfer learning approach to use. Some of these options include feature extraction, fine-tuning, and freezing specific layers while re-training other layers. During this experiment, we will use transfer learning to perfect the last few layers of the models by making adjustments to their parameters. In order to do this, the weights of the earliest layers of the pre-trained network will be maintained while the last few layers of the network will have their parameters adjusted.

\subsubsection{Fine-tuning}\hfill 

Fine-tuning is the process of tweaking the value of the model's hyper-parameters in order to get a good model. 
The variables that are established before to the beginning of the training process are referred to as hyper-parameters. When it comes to achieving the best possible performance from a network, hyper-parameters play a crucial role. As a consequence of this, we need to modify these hyper-parameters in order to improve the performance of our models. The learning rate, the number of epochs, the batch size, and the choice of activation functions are some of the important hyper-parameters that will be modified for these study.

\section{Results and discussion}
In this section, the results and discussions of the research are presented and analyzed. The chapter begins with an overview of the experimental setup, including the configuration of the models and evaluation metrics used. The results obtained from applying the selected methodologies, including various CNN architectures and vision transformers, to the cancer dataset are then thoroughly examined and discussed. The findings related to tumor detection, subtype classification, and prognosis prediction are highlighted and compared, shedding light on the performance and effectiveness of each approach. Additionally, any unexpected or interesting findings are explored, offering potential insights for future research. The chapter concludes by summarizing the main findings, addressing limitations, and suggesting further directions for exploration in the field of cancer classification and detection. 
\subsection{Residual networks}

The capacity of ResNet to model complicated connections and capture nuanced patterns from medical images makes it a valuable tool for breast cancer categorization \cite{9216455,baccouche2022integrated,article1}. ResNet's  deep architecture and skip connections lead to precise tumor categorization, which in turn facilitates earlier diagnosis and more tailored treatment plans for patients with breast cancer. In this study we use a ResNet with $ 34 $ convolutional layers. The composition of the architecture is described in figure \ref{Resn}. 
\subsubsection{Model description and advantage}\hfill

\begin{figure}[h!]
	\centering
	\includegraphics[width=0.7\linewidth]{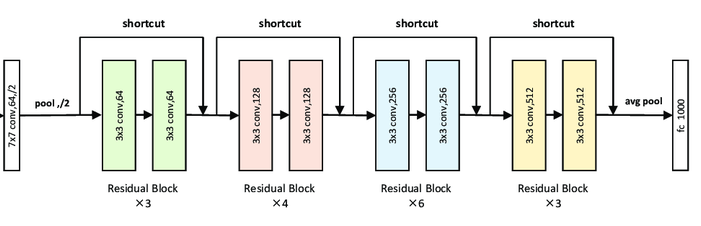}
	\caption[Structure of Resnet 34]{Structure of Resnet 34 \cite{article}.}
	\label{fig:resnet-architectures-34-101}
\end{figure}

We consider here a resnet model with $ 34 $ convolutional layers (Figure \ref{fig:resnet-architectures-34-101}) as it has not yet been studied for breast cancer classification. The first layer is made of a convolutional layer with 64 filters or kernels and a stride of 2 followed by a pooling operation. Then we have 16 residual blocks with two convolutions each. Finally, we have  a fully connected layer with a  binary output helping for our classification. In each block, we apply batch normalization after each convolution and activate with a ReLu function  (see eq. \eqref{relu}). The advantage of the ReLU function has been given in Section \ref{key} of chapter \ref{chapter3}. For the batch normalization, it helps to fasten the training procedure.

% TODO: \usepackage{graphicx} required

\subsubsection{Dataset splitting}\hfill

In this case, out of a total of $ 277,524 $ images, $ 55,506 $ were chosen for testing and $ 222,018 $ were used for training. The model was trained using around $ 80\% $ of the whole data set, while the remaining $ 20\% $ was utilized for  testing purposes.  This was done so that the test set could be kept separate (totally different from the training set). Moreover, the training phase did not make use of the data that was later used for the test set. In other words, the pictures which were used in the training phase were not employed for the testing portion of the experiment. Therefore, the effectiveness of the model was judged based on how well it prevented information from leaking out. The repartition of the positive IDC and negative IDC images is given in the Table \ref{set_spt} below. 

\begin{table}[!h]	
	\caption{Data set repartition between training, testing and validation set for ResNet}\label{set_spt}
	\begin{center}
		\begin{tabular}{|c|cc|}
			\hline
			\textbf{Cancer type}$\backslash$\textbf{Sets}	& \textbf{Training} & \textbf{Testing } \\
			\hline
			\textbf{Positive IDC }	& $63,028 $ & $ 15,758 $ \\
			\textbf{Negative IDC}	& $ 158,990  $ & $ 39,748 $ \\
			\hline
			\textbf{Total}	& \textbf{$ 222,018 $ }& \textbf{$ 55,506 $ }\\
			\hline
		\end{tabular}
	\end{center}

\end{table}
\subsubsection{Model performance}\hfill

When testing the model on the test set is achieve an amazing accuracy of $  90.40\%. $
The experiment have been done in $248708.53$ seconds, that is quite 69 hours.

We also evaluate the model with a sample of $5555$ images and got the scores in Table \ref{score}

\begin{table}[h!]	
	\caption{Performance metric on Resnet for breast cancer classification}\label{score}
	\centering
	\begin{tabular}{|c|c|c|c|}
		\hline
		\textbf{Class$\textbackslash$Score} & \textbf{Precision} & \textbf{Recall} & \textbf{F1-score} \\
		\hline
		\textbf{Positive Cancer} & $71.82\%$ & $84.46\% $ & $77.40\% $ \\
		\hline
		\textbf{Negative cancer} & $93.29\% $ & $87.35\% $ & $90.16\%$ \\
		\hline
	\end{tabular}

\end{table}
The Figure \ref{resnet}
below shows the evolution of the training loss during with the iterations. % TODO: \usepackage{graphicx} required
\begin{figure}[h!]
	\centering
	\includegraphics[width=0.4\linewidth]{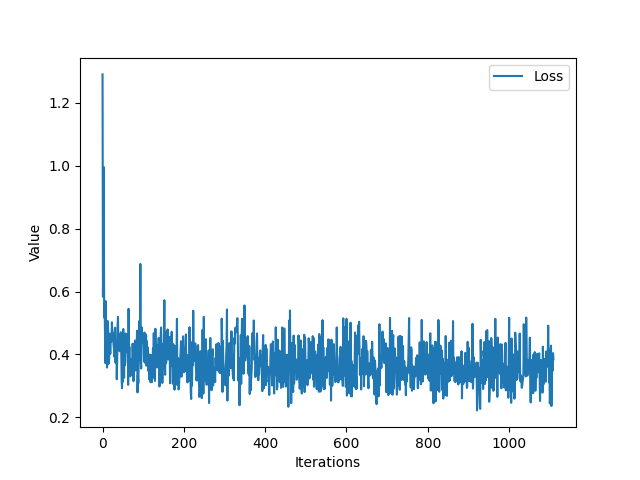}
	\caption[Evolution of Resnet loss.]{Evolution of Resnet loss.}
	\label{resnet}
\end{figure}

\subsubsection{Discussion}\hfill

The experiment with Resnet was the most accurate among the proposed models for classifying breast cancer images and achieved an accuracy score of $90.40\%$. The training was done without transfer learning so the weight values were learned from scratch. The experiment on the small sample image has shown an higher precision for the negative class which can be explained by the difference in number of the two classes images. The performance we obtained is satisfactory compare to other breast cancer classification-based models ($99.1\%$ for pretrained Resnet-50\cite{9216455}, $98.77\%$ for pretrained Resnet-101 \cite{gandomkar2018mudern}) because it has lower depth.

\subsection{Convolutional neural network}

Convolutional Neural Networks are highly effective for image classification tasks, including the classification of breast cancer images. They excel in analyzing visual data by utilizing convolutional layers to extract meaningful features from images. 

\subsubsection{Model description and advantage}\hfill

We utilize a convolutional network with 33 convolutions and 2 dense layers. This choice was motivated by the use of Resnet 34 in the other architecture to be able to compare the two. The depth of the architecture allows it to learn complicated patterns which makes it suitable for our classification. After every hidden layer, we use a ReLU activation function  because of its ability to fasten the learning process. The  dense layers are used at the end for classification and the output is activated using sigmoid. The convolutions are built with 64 filters and kernel size of $3\times3$.

\subsubsection{Dataset splitting}\hfill

We split the data into two subsets, namely a training and a testing set. The proportions are the similar to the case of Resnet (see Table \ref{set_spt}). 
We load the subsets using data loader methods from Pytorch and use batch size of 100. This method reduced considerably the running time of the model.
\subsubsection{Model evaluation and performance}\hfill
% TODO: \usepackage{graphicx} required
\begin{figure}[!h]
	\centering
	\includegraphics[width=0.4\linewidth]{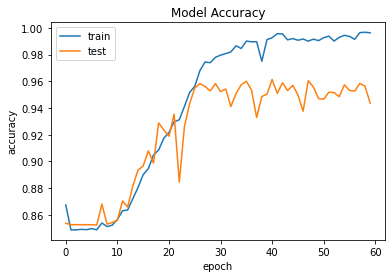}
	\includegraphics[width=0.4\linewidth]{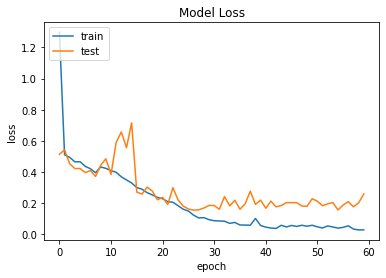}
	\caption{Accuracy an loss evolution.}
	\label{fig:accuracycnn}
\end{figure}

We used 60 epochs for training the model. The training procedure was about 6 hours of time. The model achieved an accuracy of $71.64\%. $

\subsection{Results with VGG-16}
We split the data using the ratio 70:15:15. The choice of splitting the data into $ 70\% $ for training, $ 15\% $ for validation, and $ 15\% $ for testing is a common practice in machine learning. This split provides enough data for training, enables effective hyper-parameter tuning and model selection, and allows for unbiased evaluation of the model's performance on unseen data. The training set provides a substantial amount of data for the model to learn from, the validation set aids in optimizing hyper-parameters, and the testing set ensures an unbiased assessment of the model's generalization ability. By keeping the datasets separate, the integrity of the evaluation process is maintained, and the model's real-world performance can be reliably estimated.

\begin{table}[!h]
	\caption{Data set repartition between training, testing and validation set for VGG}\label{set_split}
\begin{center}
		\begin{tabular}{|c|ccc|}
	\hline
\textbf{Cancer type}$\backslash$\textbf{Sets}	& \textbf{Training} & \textbf{Testing }& \textbf{Validation} \\
	\hline
\textbf{Positive IDC }	& $55,150 $ & $ 11,818 $ & $ 11,818 $ \\
\textbf{Negative IDC}	& $ 139,116  $ & $ 29,811 $ & $ 29,811 $ \\
	\hline
\textbf{Total}	& \textbf{$ 194,266 $ }& \textbf{$ 41,699 $ }& \textbf{$ 41,699 $} \\
\hline
\end{tabular}
\end{center}

\end{table}

\subsubsection{Experimental condition}\hfill

The stochastic gradient descent algorithm, with a learning rate of 0.001, was employed as the optimizer for the system. The model was tweaked to perfection using the procedures outlined in Section \ref{fine}.  For the purpose of fine-tuning, the first 15 convolutional layers of the VGG-16 model were frozen. The stochastic gradient descent optimizer was then used to train the last convolutional layer as well as the fully connected layers. 

\subsubsection{Model evaluation and performance}\hfill

\begin{figure}[h!]
	\centering
		\includegraphics[width=0.45\linewidth]{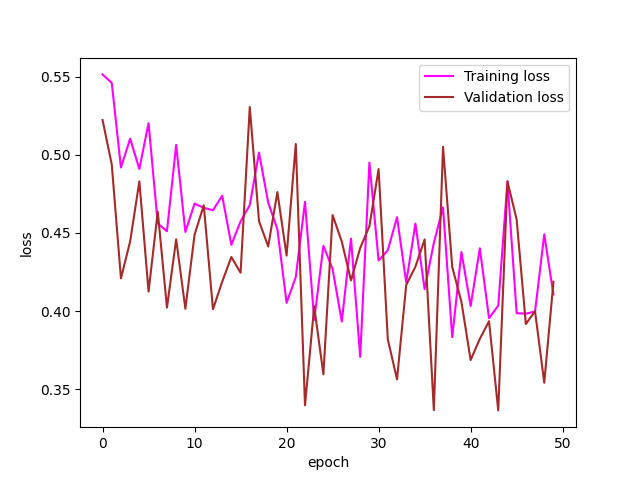}\label{rev}
	\caption[Training and validation loss and accuracy.]{Training and validation loss and accuracy.}
	\label{fig:vggaccuracyevol}
\end{figure}

For the training part we use a total of 50 epochs and the weight were trained using stochastic gradient descent. The evolution of the loss and accuracy during training are given in Figure \ref{fig:vggaccuracyevol}. The model training finishes after 49 hours and achieve accuracy score of $ 86.81\% $ . The precision and F1-score are given in Table \ref{scores} below.

\begin{figure}[h!]
	\centering
	\includegraphics[width=0.45\linewidth]{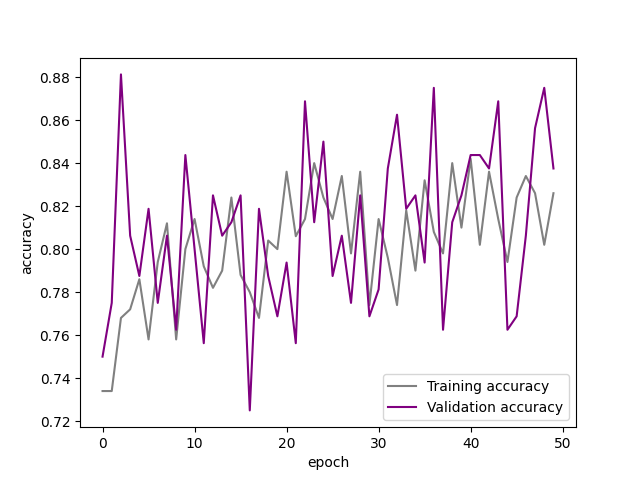}\label{revv}
	\caption[Training and validation loss and accuracy.]{Training and validation loss and accuracy.}
	\label{figv:vggaccuracyevolv}
\end{figure}

\begin{table}[h!]
	\caption{Performance metric on VGG for breast cancer classification}\label{scores}
	\centering
	\begin{tabular}{|c|c|c|c|}
	\hline
	\textbf{Class$\textbackslash$Score} & \textbf{Precision} & \textbf{Recall} & \textbf{F1-score} \\
	\hline
	\textbf{Positive Cancer} & $78\%$ & $75\% $ & $76\% $ \\
	\hline
	\textbf{Negative cancer} & $90\% $ & $92\% $ & $91\% $ \\
	\hline
\end{tabular}

\end{table}

\subsection{Results with vision transformer}

Vision transformers have several advantages for breast cancer classification, including improved accuracy compared to other models, reduced computational complexity, effective transfer learning capabilities, and flexibility in handling different types of breast cancer imaging data. These benefits make vision transformers a promising approach for enhancing breast cancer classification accuracy while maintaining computational efficiency and adaptability to diverse imaging datasets.

\subsubsection{Model description and experimentation}\hfill

The understanding of how Visions Transformers works can be seen in Section \ref{vit}.
The data set is split the same way like for ResNet (table \ref{set_spt}). Table \ref{ex_condit} displays the different parameters in our model an their value. 

\begin{table}[!h]
		\caption{Experimental condition for training}\label{ex_condit}
	\begin{center}
		\begin{tabular}{|c|c|}
			\hline
			\textbf{Parameters}&\textbf{Values}	\\ 
			\hline
			\textbf{Number of epochs }	& $10$ \\
			\textbf{Learning rate}	& $ 0.001 $ \\
			\textbf{Weight decay}	& $ 0.0001 $ \\
			\textbf{Batch size}	& $ 10 $ \\
			\textbf{Dimension of projection}	& $ 32 $ \\
			\textbf{Number of heads in the encoder}	& $ 8 $ \\
			\textbf{New image size}	& $ 72 $ \\
			\textbf{Patch size}	& $ 6 $ \\
			\textbf{Loss function }	& sparse categorical Cross-entropy loss from Keras \\
			\textbf{Optimizer}	& Adam from tensorflow-addons \\
			\hline
		\end{tabular}
	\end{center}

\end{table}
The image new size is set to $72\times72$ due to computational and time limitation. However, there is enough patches to capture attention in each images. The batch size is 10 because in the experimentation it helps the learning to happen faster. 
\subsubsection{Model evaluation and performance}\hfill

The training was achieved in 24 hours on the whole $80\%$ of the dataset. The evaluation of the model on unseen data displayed an accuracy of $83.33\%$. We also use a sample of $3331$ images to access the precision, recall and F1-score were the results are given in Table \ref{scor}.

\begin{figure}[h!]
	\centering
	\includegraphics[width=0.7\linewidth]{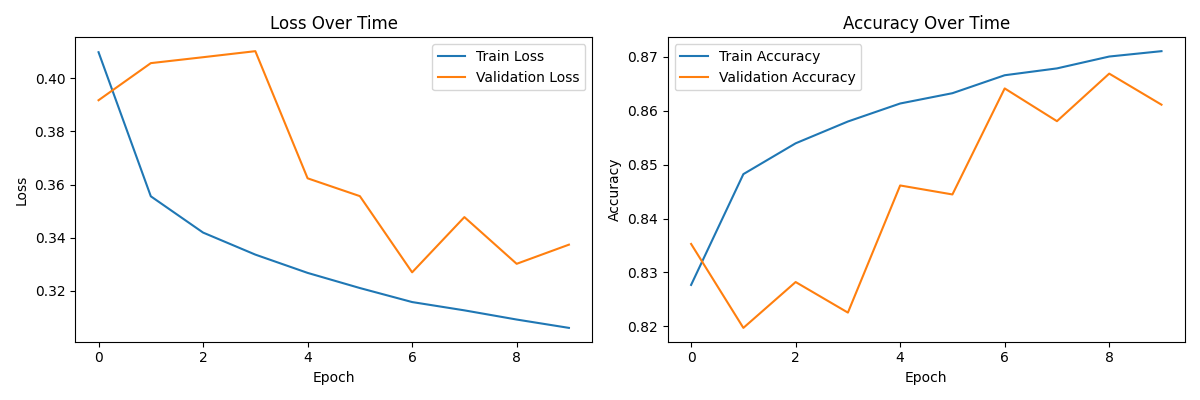}
	\caption[Accuracy and loss evolution with ViT.]{Accuracy and loss evolution with ViT.}
	\label{fig:vitransplots}
\end{figure}

\begin{table}[h!]	
	\caption{Performance metric for  on ViT for breast cancer classification}\label{scor}
	\centering
	\begin{tabular}{|c|c|c|c|}
		\hline
		\textbf{Class$\textbackslash$Score} & \textbf{Precision} & \textbf{Recall} & \textbf{F1-score} \\
		\hline
		\textbf{Positive Cancer} & $62.83\%$ & $84.12\% $ & $72.20\% $ \\
		\hline
		\textbf{Negative cancer} & $93.25\% $ & $82.44\% $ & $87.31\% $ \\
		\hline
	\end{tabular}

\end{table}

\subsection{Discussion} 
In this section, we discuss the result we have found previously. 
\begin{itemize}
\item 	The two graphs Figure \ref{figv:vggaccuracyevolv} and Figure \ref{fig:vggaccuracyevol} show the training loss and testing loss for VGG-16 and Vision transformers respectively. The first graph (Figure \ref{figv:vggaccuracyevolv}) shows that VGG-16 is not overfitting the training data, while the second graph (Figure \ref{fig:vggaccuracyevol}) shows  that the complexity of vision transformer  architecture makes it overfit the training data. This is because the Vision Transformers has more parameters to learn, and as a result, it takes longer for the model to converge. The ideal situation would be for the training loss and testing loss to decrease together, but this is not always possible. In this case, the goal is to minimize the testing loss as much as possible and one can do that by reducing the number of parameters of the model.

Overall, the Figure \ref{fig:vggaccuracyevol} shows that the VGG-16  model is learning effectively, while Figure \ref{rev} shows that the ViT is overfitting the training data.

\item Resnet-34 appears to be learning effectively as we can see in Figure \ref{resnet}. This raises the model to the best model in our experiments as it exhibits the best accuracy among the proposed models. However, the power of transfer learning heights VGG-16 in the first position when analyzing F1-score, which for this case is $91\%$ for the negative class and $76\%$ for the positive class. This is the best model overall since we are dealing with imbalanced data. Moreover, it finishes after 49 hours. 
\item The running time analysis of the models shows that the vision transformer model is fast in learning but it does not perform as well as Resnet and VGG. However, we will notice that the experiments have been done in different conditions. While Resnet took 69 hours to run using 50 epochs, ViT took 24 hours to complete using 10 epochs. The VGG-16 model due to transfer learning showed a rapid training process and finished in 49 hours for 50 epochs.
\item  The CNN model showed the least accuracy measure. This is because the architecture used same convolutional layers stacked together reducing the learning ability of the model. 

\item  The recall and precision recorded from the different models show the imbalance in the data. The fact that the negative class is associated with highest values of these measure is due to the higher number of negative class. Resnet and ViT demonstrate a great ability to recognize negative class images with precision of $93\%$. VGG-16 performed also well with a precision of $90\%$.

\item Resnet has the highest F1-score for the positive class followed by VGG-16. This result demonstrate the power of Resnet and VGG to handle imbalanced data (Table \ref{scores}).
\end{itemize}

The python codes used to generate all the results can be found \href{https://github.com/Jordan-buwa/planets/tree/Breast_cancer_classification}{here}.

\section{Conclusion}

In this research, we provided a detailed study on the use of deep transfer learning for the classification of breast cancer. We employed a Vision Transformers, a Resnet-34, a plain CNN and a VGG-16 to classify images of invasive ductal carcinoma cancer. The result has demonstrated the power of residual networks in image recognition putting Resnet-34 at the top with an accuracy of $90\%$. The network structure with skip connection allowed an effective learning of image features. However, due to the imbalanced data, we shall rely on the F1-score measure  for which VGG outperformed. We noticed that the use of transfer learning is required for fast training and better generalization of models . Vision transformer demonstrated its power in classifying images. However, the model experienced overfitting due to the multitude of parameters it takes in account. The VGG-16 showed a good learning trend due to the fact that the model has been pretrained and therefore recognize some feature already which makes it to update quickly the weights of the network and minimize the error in predictions.

 We did not perform all the experiments in uniform experimental conditions by the fact of unavailability of strong computational resources and time limitation. So, we exhibit some plan for future works:
 \begin{itemize}
 	\item Investigate the use of a new loss function to improve the accuracy of the models.
 	\item Perform detection and  Multi-class classification of breast cancer abnormalities.
 \end{itemize}

\paragraph{Acknowledgments}
			This work was supported by Michigan State University and AIMS Ghana.

\newpage

\bibliography{biblio_jmmbiblio_jmm}
\bibliographystyle{plain}
\end{document}